\documentclass{styles/svproc}
\usepackage{url}
\usepackage{graphicx}
\usepackage{float}
\usepackage{hyperref}
\usepackage{caption}
\usepackage{subcaption}
\usepackage{diagbox}
\usepackage{rotating}
\usepackage{amsmath}

\begin{document}
\mainmatter              
\title{Visual Context-Aware Person Fall Detection}
\titlerunning{Visual Context-Aware Person Fall Detection}  
%
\author{Aleksander Nagaj\inst{1}\inst{2} \and Zenjie Li\inst{2} 
\and Dim P. Papadopoulos \inst{1} \and Kamal Nasrollahi\inst{2}\inst{3}}
\authorrunning{Aleksander Nagaj et al.} 
%
\tocauthor{Aleksander Nagaj, Zenjie Li, and Dimitrios Papadopoulos}
\institute{Technical University of Denmark, DTU, Anker Engelunds Vej 101,\\2800 Kongens Lyngby, Denmark\\
\and
Milestone Systems A/S, Banemarksvej 50 C, 2605 Brøndby, Denmark
\and
Aalborg University, Fredrik Bajers Vej 7K, 9220 Aalborg
}

\maketitle              

\begin{abstract}
As the global population ages, the number of fall-related incidents is on the rise. Effective fall detection systems, specifically in healthcare sector, are crucial to mitigate the risks associated with such events. This study evaluates the role of visual context, including background objects, on the accuracy of fall detection classifiers. We present a segmentation pipeline to semi-automatically separate individuals and objects in images. Well-established models like ResNet-18, EfficientNetV2-S, and Swin-Small are trained and evaluated. During training, pixel-based transformations are applied to segmented objects, and the models are then evaluated on raw images without segmentation. Our findings highlight the significant influence of visual context on fall detection. The application of Gaussian blur to the image background notably improves the performance and generalization capabilities of all models. Background objects such as beds, chairs, or wheelchairs can challenge fall detection systems, leading to false positive alarms. However, we demonstrate that object-specific contextual transformations during training effectively mitigate this challenge. Further analysis using saliency maps supports our observation that visual context is crucial in classification tasks. We create both dataset processing API and segmentation pipeline, available at \href{https://github.com/A-NGJ/image-segmentation-cli}{https://github.com/A-NGJ/image-segmentation-cli}.
\keywords{fall detection, visual context, computer vision, data augmentation}
\end{abstract}
%


\begin{figure}[htb]
    \centering
    \begin{subfigure}{\textwidth}
        \centering
        \includegraphics[width=\linewidth]{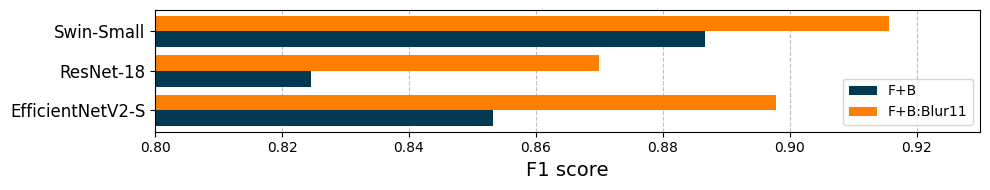}
        \caption{}
        \label{fig:enter-label}
    \end{subfigure}
    \begin{subfigure}{.49\textwidth}
        \centering
        \includegraphics[width=.5\linewidth]{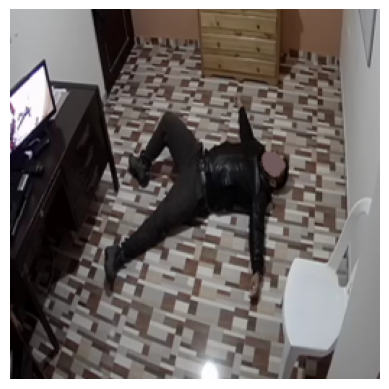}
        \caption{}
        \label{subfig:raw-image}
    \end{subfigure}
    \begin{subfigure}{.49\textwidth}
        \centering
        \includegraphics[width=.5\linewidth]{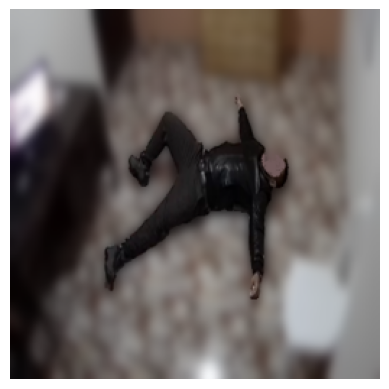}
        \caption{}
        \label{subfig:background-blur}
    \end{subfigure}
    \caption{Comparison of the performance of models trained on raw Foreground+Background (F+B) data (see \autoref{subfig:raw-image} as an example), and transformed data with background smoothed using Gaussian blur with kernel size 11 (F+B:Blur11) (see \autoref{subfig:background-blur} as an example). All models were tested on raw images without segmentation. The F1 score is increased for each architecture when training on transformed data. This image is sourced from the CAUCAFall dataset \cite{eraso2022CAUCAFall}.}
    \label{fig:main-figure}
\end{figure}

\setcounter{footnote}{0}
\section{Introduction}

According to the World Health Organization\footnote{https://www.who.int/news-room/fact-sheets/detail/falls}, over 37 million severe falls requiring medical attention occur worldwide annually. A fall detection system can effectively mitigate the risk associated with a fall accident by allowing for a faster response time. One of the predominant strategies involves the use of sensor-based approaches, including accelerometers or other wearable sensors\footnote{https://www.lifeline.com/medical-alert-systems/fall-detection/}\footnote{https://support.apple.com/en-us/108896}\footnote{https://www.medicalguardian.com/}. These devices must be worn constantly, causing discomfort over time and often requiring frequent charging. On the contrary, emerging camera-based Computer Vision (CV) systems do not require physical interaction and rely purely on captured videos or images. With the advancements in Deep Learning (DL), including robust Convolutional Neural Networks (CNN) and Visual Transformers (ViT) \cite{dosovitskiy2021image}, vision-based fall detectors can be non-intrusive and applicable at scale using multi-camera setups in, i.e., hospitals or nursing homes \cite{Madsen2024}.

The accuracy of a fall detector depends greatly on the available training data and operating environment. Each scene contains objects that can affect detection accuracy. Objects not central to an analysis but appearing in an image are referred as the \textit{Visual Context} in this study. We integrate segmentation and transformation techniques to manipulate visual context and thoroughly evaluate our methods using well-established models such as ResNet18 \cite{he2015deep}, EfficientNetV2-S \cite{tan2019efficientnet}, and Swin-Small \cite{liu2021swin}. Leveraging Segment Anything Models (SAM) \cite{kirillov2023segment} and saliency maps, our research provides insight into how scene elements affect detection accuracy.

The contributions of this work are multifold:
\vspace*{-6.3pt}
\begin{itemize}
    \item Enhancing fall detection performance by visual context-aware augmentation during training.
    \item Advancing understanding of the role of visual context in fall detection, paving the way for more reliable systems.
    \item Creating a fall dataset with annotated visual context based on public datasets. Furthermore, we create a Python API that facilitates the usage of our or any other dataset annotated with our segmentation pipeline.
\end{itemize}
More specifically, we show that Gaussian blur applied to the background, i.e., everything except the subject, evidently improves the performance of all evaluated model architectures (\autoref{fig:main-figure}).

\section{Related Work}

\subsection{Role of visual context in computer vision}

\textit{Moyaeri et al.} \cite{moayeri2022comprehensive} investigated image classification models, finding that adversarial training enhances the background sensitivity of ResNets, while contrastive training diminishes foreground sensitivity. They also noted the adaptability of ViT architecture models in increasing foreground sensitivity with rising image corruption. \textit{Xiao et al.} \cite{xiao2020noise} analyzed the reliance of object recognition models on background signals, indicating that higher accuracy models tend to rely less on background information. Both studies, however, relied on manual segmentation methods prone to inaccuracies, focusing mainly on foreground-background separation without exploring more detailed semantic segmentation. Addressing these limitations, our research employs advanced semantic segmentation techniques for a more refined analysis, improving comprehension of complex visual contexts.

\subsection{Advances in CV foundation models}

\textit{Kirillov et al.} \cite{kirillov2023segment} from Meta AI Research introduced the Segment Anything Model (SAM). It showcases exceptional generalization capabilities and flexibility in segmentation prompts, making it ideal for complex downstream tasks. \textit{Liu et al.} \cite{liu2023grounding} developed GroundingDINO, an open-set object detector that generates bounding boxes for objects specified by a text prompt. Our research leverages both SAM and GroundingDINO to enhance object annotation in the fall dataset, demonstrating how the text-prompt guided object detection combined with advanced semantic segmentation can provide a robust labeling tool.

\section{Dataset}

There exist a few publicly available fall datasets. URFall \cite{kepski2014falldataset} contains 70 image sequences, divided into 30 fall events and 40 activities of daily living (ADL) recorded with Microsoft Kinect cameras in an office and a private house. KULeuven \cite{baldewijns2016bridging} includes 55 fall and 17 ADL scenarios, all recorded with 5 cameras in a setting designed to resemble a nursing home. CAUCAFall \cite{eraso2022CAUCAFall} was captured with one camera in a home environment where each of 10 participants performed 5 types of falls and 5 ADLs. Despite all subjects being in a relatively young age and all datasets depicting a rather limited environment, they are a valuable source of data that can be used for training and evaluation of CV-based fall detectors.

\subsection{Our dataset with semantically segmented objects}

In the development of the dataset for visual context analysis in fall detection, we extracted frames from the previously public datasets. CAUCAFall offered 7,388 fall and 12,466 non-fall images, which, after temporal downsampling and removal of ambiguous frames, yielded 1,538 fall and 1,575 non-fall images. For KULeuven, from 55 fall videos, we extracted frames at 0.5 fps for ADL and 2fps for fall events, resulting in 713 fall and 1,950 non-fall images. We chose more frequent frame rate for falls to capture their more dynamic nature. From URFall, 53 fall and 323 non-fall images from RGB Camera 0 were distilled to 42 fall and 275 non-fall images after excluding ambiguous content. Person detection via a pre-trained YOLOx \cite{yolox2021} model, with bounding boxes doubled in size, captured the subject and surroundings. Images were then cropped to the bounding box area. Next, using the segmentation pipeline detailed in \autoref{sec:methods}, images were segmented, annotated, and aligned to a standardized format for consistency. \autoref{tab:dataset-stats} presents a final share of each public dataset in our dataset.


\begin{table}[htb]
    \centering
    \begin{tabular}{l|r|r|r}
        \textbf{Name} & \textbf{No. falls} &\textbf{ No. non-falls} & \textbf{Falls + non-falls} \\
        \hline
        CAUCAFall \cite{eraso2022CAUCAFall} & 1,538 & 1,575 & 3,113 \\
        KULeuven \cite{baldewijns2016bridging} & 713 & 1,950 & 2,663  \\
        URFall \cite{kepski2014falldataset} & 42 & 275  & 317  \\
        \hline
        Sum & 2,293 & 3,800 & 6,093 
    \end{tabular}
    \caption{Share of fall and non-fall images across subsets in our merged dataset.}
    \label{tab:dataset-stats}
\end{table}

To ensure minimal information leakage between training and testing phases, the dataset was strategically divided into training and test subsets, presented in \autoref{tab:training-test-set}. The test set contains images from URFall, a dataset with notably inferior image quality, and from three out of five KULeuven cameras, thereby introducing a varied testing scenario. The training set consists of the remaining images. KULeuven, with its multi-camera setup facilitated evaluation of models' capability to accurately interpret identical scenes captured from multiple angles. Additionally, the test set was intentionally imbalanced, with a lower representation of falls, to mirror real-world scenarios, where fall are less prevalent than ADL.

\vspace*{-25pt}
\begin{table}[htb]
    \centering
    \begin{tabular}{l|l|r|r}
    \textbf{Subset} & \textbf{Source} & \textbf{Fall} & \textbf{Non-fall} \\
    \hline
    Training & CAUCAFall & 1538 & 1575 \\
    Training & KULeuven cameras 3, 4, and 5 & 429 & 1174 \\
    Test & KULeuven cameras 1 and 2 & 284 & 776 \\
    Test & URFall & 42 & 275 \\
    \end{tabular}
\caption{Training and test set split of our fall dataset.}
\label{tab:training-test-set}
\end{table}

\section{Methods}
\label{sec:methods}

\subsection{Segmentation techniques}

\begin{figure}[tb]
    \centering
    \includegraphics[width=0.9\textwidth]{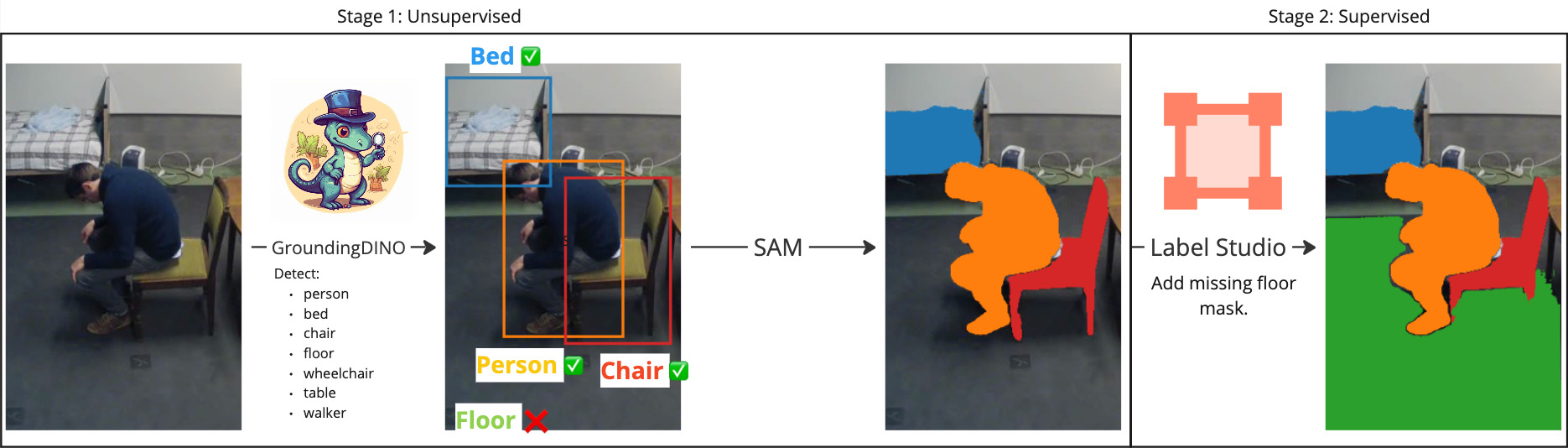}
    \caption{Labeling pipeline. In the first, unsupervised stage, GroundingDINO produces bounding boxes for detected objects that are used as input prompts for SAM, resulting in segmented objects with semantic text labels. In the second, supervised stage, produced annotations are inspected and refined in Label Studio.}
    \label{fig:segmentation-pipeline}
\end{figure}

In the following, we describe the proposed annotation pipeline comprising GroundingDINO, SAM, and Label Studio\footnote{https://labelstud.io/} for semantic image segmentation (\autoref{fig:segmentation-pipeline}). The process is divided into two stages. In the first, unsupervised stage, GroundingDINO identifies key objects and produces bounding boxes for each object that are then used as input prompts for SAM which performs semantic segmentation. In this stage, we define a person, a chair, a table, a bed, a wheelchair, floor, and a walking aid (walker) as key objects. Furniture that serves a similar purpose is considered as one of the previously mentioned labels for simplification. The binary masks generated by SAM are stored in a common COCO JSON format\footnote{https://cocodataset.org/\#formatdata}. In the second, supervised stage, masks are enhanced in Label Studio, utilizing a custom interface backed by SAM for thorough label verification and adjustment. This dual-phase approach ensures high-quality segmentation masks encoded in Run-Length Encoding (RLE) format.

\subsection{Contextual transformations}
\label{subsec:contextual-transformations}

\begin{figure}[tb]
    \captionsetup{font=footnotesize}
    \centering
    \begin{subfigure}{.32\textwidth}
        \centering
        \includegraphics[width=.7\linewidth]{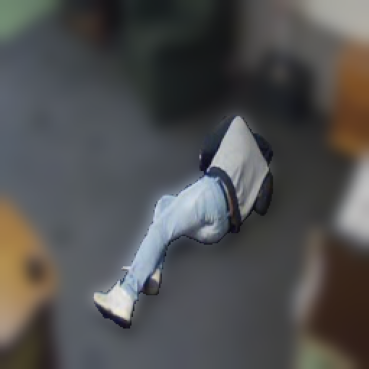}
        \caption{Background (person inverse): Gaussian blur.}
        \label{sub:transformations-raw-image}
    \end{subfigure}
    \begin{subfigure}{.32\textwidth}
        \centering
        \includegraphics[width=.7\linewidth]{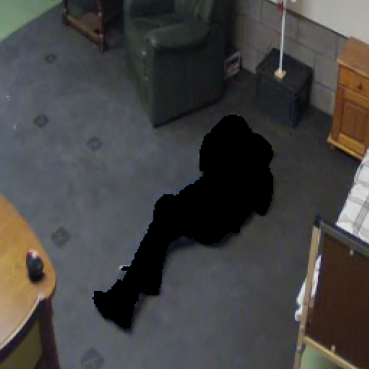}
        \caption{Person (foreground): solid black.}
        \label{sub:transformations-background-blur}
    \end{subfigure}
    \begin{subfigure}{.32\textwidth}
        \centering
        \includegraphics[width=.7\linewidth]{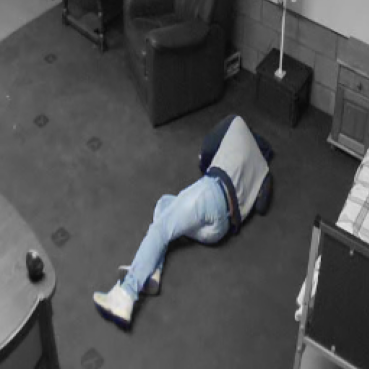}
        \caption{Background: grayscale.\newline}
        \label{sub:transformations-person-plur}
    \end{subfigure}
    \captionsetup{font=small}
    \caption{Pixel-based transformation applied to different areas of an image.}
    \label{fig:augmentation-methods}
\end{figure}

Our study focuses on modifying contextual information to gain meaningful insights into how visual context affects fall detectors. We focus on a data-centric approach, applying pixel-based transformation methods to the labeled dataset before model training. Transformations include: \textbf{solid color} (to assess model reliance on background information), \textbf{Gaussian blur} (to test model sensitivity to detail clarity), and \textbf{grayscale} (to evaluate the importance of color information). \autoref{fig:augmentation-methods} presents examples of transformations applied to one or more key objects. First, a transformation is applied to the entire image, then binary mask representing one or more key objects is applied, leaving the remaining parts of the image intact. Additionally, we apply augmentation methods in the following order: resize to $256\times256$, random perspective with distortion scale 0.4 (training only), random horizontal flip (training only), transform to PyTorch tensor, normalize. Those augmentation methods are applied \textit{after} contextual transformations except Gaussian blur, which is applied either before or after resizing image to a fixed resolution. In this way, we have devised two scenarios: one involving controlled smoothing (image resizing before applying image blur) and another featuring a seeded kernel size (image resizing after applying image blur). 

In our evaluation, we utilize the F1 score as the primary metric. This metric is particularly suitable for assessing imbalanced datasets, which is typical in fall detection tasks where actual fall events are infrequent compared to normal activities.

\section{Experiments}

Each model was trained with the following hyperparameters: batch size 32, learning rate 0.001, momentum 0.9, 50 epochs, and early stopping on validation loss with patience 5. To mitigate the risk of having nearly identical frames in both validation and training sets, images were initially grouped into sets of five consecutive frames. These groups were then randomly split, maintaining a 10\% allocation for the validation set.

\subsection{Nomenclature}

In this context, ``Foreground'' (F) refers to the main object of interest, typically a person, while ``Background'' (B) encompasses everything outside the foreground, essentially the inverse of the foreground. The nomenclature for test scenarios is as follows:

{\footnotesize
\begin{align*}
    & F:\text{[transform]}+ B:\text{[transform]} \\
\end{align*}
}
Square brackets [] denote optional parameters. Transformation applies to one or or key objects, or their inverse.

\subsection{Effect of Gaussian blur transformation}

To implement a Gaussian blur transformation, there is a need to specify a kernel size. We train each model on images with background transformed with Gaussian blur with increasing kernel size F+B:Blur$x$, adhering to condition $x = 2p + 1, p \in \mathcal{N}, 3 \leq x \leq 31$. We employ two Gaussian blur strategies as described in \autoref{subsec:contextual-transformations}. Models were evaluated on the test set F+B, without contextual transformations applied, unless otherwise stated.

\begin{figure}[tb]
    \centering
    \includegraphics[width=0.85\textwidth]{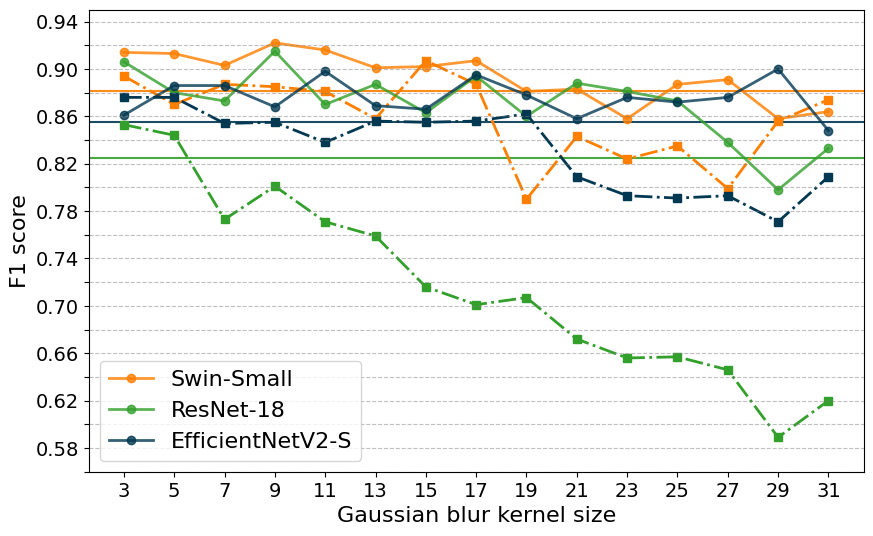}
    \caption{F1 scores over Gaussian blur kernel size for different model architectures. Horizontal lines represent the F1 score for baseline F+B model (trained without contextual transformations). Solid lines with dot markers represent models trained with seeded kernel (blur before resize). Dashdot lines with square markers represent models trained with a fixed kernel (blur after resize).}
    \label{fig:gaussian-blur-kernel-size}
\end{figure}

As shown in \autoref{fig:gaussian-blur-kernel-size}, a seeded kernel, which allows for some degree of randomness in smoothing strength, results in significantly more robust results than a fixed kernel. Moreover, a modest kernel size often performs best. Without jeopardizing the generality of the study, we proceeded with a seeded kernel Gaussian blur with a kernel size of 11 in further experiments.

\subsection{Understanding the influence of visual context}

To understand the impact of visual context on fall detector performance, we trained each model on transformations (see \autoref{subsec:contextual-transformations}), applied to the foreground or background. \autoref{tab:understanding-influence-of-visual-context} presents evaluation results on F+B test set. We observe a marked robustness to blur transformations, where training on images with blurred background boosts performance on raw F+B data, suggesting that smoothing supports more generalizable feature learning. Conversely, drastic alternations, like a solid black transformation, degrade performance and lead to overfitting. CNNs are more reliant on color information that Transformers as grayscale transformation causes a significant drop in performance of both ResNet-18 and EfficientNetV2-S. This analysis highlights the dominance of foreground information over background but stresses the importance of both for optimal visual data interpretation, showcasing their complex interplay in classification tasks.

\begin{table}[htb]
    \hspace{-1.5cm}
    \begin{center}
    \resizebox{\linewidth}{!}{%
    \footnotesize
    \begin{tabular}{|p{2.4cm}|c|c|c|c|c|c|c|}
    \hline
        \diagbox{Test}{Train} & F+B & F+B:Blur11 & F+B:SolidBlack & F+B:Grayscale & F:Blur11+B & F:SolidBlack+B & F:Grayscale+B \\
        \hline
        F+B ResNet-18 & 0.825 & \textbf{0.87} (0.919)  & 0.551 (0.912) & 0.724 (0.871) & 0.807 (0.898) & 0.638 (0.867) & 0.722 (0.854)\\
        \hline
        F+B EfficientNetV2-S & 0.853 & \textbf{0.898} (0.83)  & 0.471 (0.872) & 0.777 (0.87) &0.825 (0.842) & 0.566 (0.824) & 0.787 (0.841)\\
        \hline
        F+B Swin-Small & 0.887 & 0.916 (0.916) & 0.732 (0.839) & 0.913 (0.895) &\textbf{0.929} (0.873) & 0.743 (0.872) & 0.926 (0.904)\\
        \hline
    \end{tabular}
    }
    \end{center}
    \caption{F1 Scores for the F+B test scenario. \textbf{Bold} indicates the best test for a model. The F1 scores in parentheses are evaluations on the test set with the same contextual transformations as in model training.}
    \label{tab:understanding-influence-of-visual-context}
\end{table}
\vspace*{-15pt}

\subsection{Significance of specific objects}

\begin{figure}[tb]
    \centering
    \includegraphics[width=\textwidth]{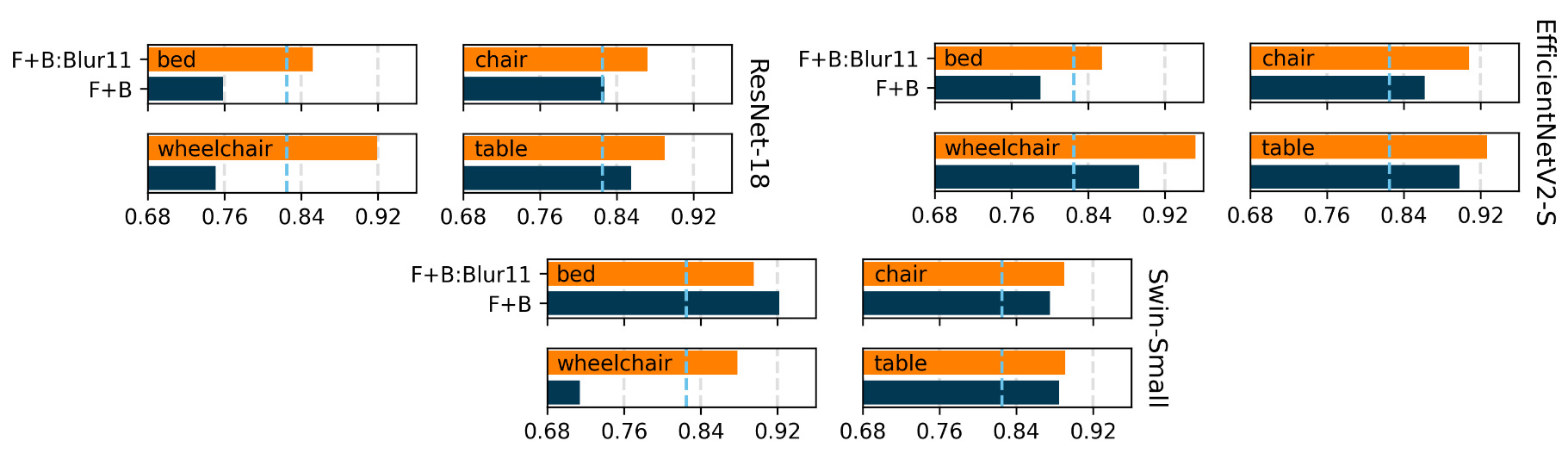}
    \caption{F1 scores for subsets of the training set containing the specific key object for each analyzed model architecture. The light blue dashed line shows the F1 score of an F+B trained model for each architecture, tested on the entire dataset.}
    \label{fig:objects-in-detection}
\end{figure}

We identify key objects as those often found in misclassified images, suggesting a model's difficulty in correctly classifying a fall when these objects are present. We assessed F+B and F+B:Blur11 models on test subsets with key objects for each architecture. Across all models, there was a notable response to the presence of key objects (\autoref{fig:objects-in-detection}). Wheelchairs posed a significant challenge for ResNet-18 and Swin-Small, whilst beds were particularly problematic for ResNet-18 and EfficientNetV2-S. People interacting with beds or wheelchairs may assume poses similar to a fall, potentially confusing a fall detector. Smoothing the background has proven effective in reducing such impact, suggesting that a less detailed background enables the fall detector to better focus on the primary subject, thereby improving classification accuracy.

\subsection{Qualitative insights from saliency maps}

\begin{figure}[tb]
    \centering
    \includegraphics[width=0.7\textwidth]{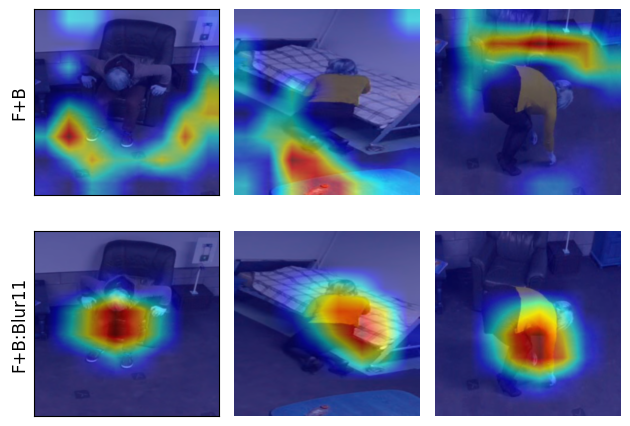}
    \caption{Saliency map examples from GradCAM for the last convolutional layer of EfficientNetV2-S. Top row: F+B models, bottom row: F+B:Blur11 models. The heatmap scale ranges from blue (lowest) to red (highest) influence.}
    \label{fig:saliency-maps}
\end{figure}

GradCAM \cite{selvaraju2017grad} is commonly utilized to visually explain CNN-based models by producing a heatmap using gradients from the selected convolutional layer, typically the final one. We explored how attention changes when training the model on a smoothed background. Due to implementation limitations, saliency maps in our study are created only for CNNs. We used default GradCAM parameters and applied them to the last convolutional layer. As shown in \autoref{fig:saliency-maps}, the model tends to shift its attention towards the person when trained on the transformed visual context, indicating improved subject distinction through background blurring.

\section{Conclusions}

Visual context is highly relevant for deep learning-based fall detection classifiers. Objects such as wheelchairs and beds tend to have strong influence on their performance. Properly applied context-aware transformations to the training data benefit the accuracy and generalization capabilities of fall detectors, leading to more reliable systems. Further effort is required to comprehend the extent and working mechanism of the impact of visual context and its applicability to other computer vision tasks. Additionally, expanding datasets to encompass diverse conditions and addressing ethical concerns are essential for advancing fall detection technologies.

%
%

\bibliographystyle{styles/bibtex/spmpsci.bst}
\bibliography{bibliography}

\end{document}